# Automatic Identification of Retinal Arteries and Veins in Fundus Images using Local Binary Patterns

Nima Hatami[a,*], Michael H. Goldbaum[a]

[a]*Shiley Eye Center, University of California San Diego, 9500 Gilman Drive La Jolla, CA 92093-0946*

## Abstract

Artery and vein (AV) classification of retinal images is a key to necessary tasks, such as automated measurement of arteriolar-to-venular diameter ratio (AVR). This paper comprehensively reviews the state-of-the art in AV classification methods. To improve on previous methods, a new Local Binary Pattern-based method (LBP) is proposed. Beside its simplicity, LBP is robust against low contrast and low quality fundus images; and it helps the process by including additional AV texture and shape information. Experimental results compare the performance of the new method with the state-of-the art; and also methods with different feature extraction and classification schemas.

*Keywords:* Computer-Aided Diagnosis (CAD), Medical image analysis, Vessel Classification, Arteries and Veins, Retinal fundus images, Arteriolar-to-venular diameter ratio (AVR), Feature extraction, Local Binary Patterns (LBP).

## 1. Introduction

Retinal vessels provide useful information for clinical diagnosis and treatment. Disease and abnormalities such as high blood pressure, branch artery occlusion, central retinal artery occlusion, branch vein occlusion, central retinal vein occlusion, diabetes, retinal vasculitis and retinal hemangioma can be predicted or observed by retinal vasculature images. Quantitative analysis of the retinal arteries and veins could help monitor disease and the effect of treatment during clinic visits. Relevant changes in arteries and veins can be subtle but are limited to qualitative analysis by ophthalmologists during clinical examination. The process of manually or semi-automatically classifying retinal vessels from digital fundus images is laborious. As part of an automatic procedure for retinal vessel analysis, it would be valuable to be able to discriminate arteries from veins.

*Corresponding author Tel: +1 (858) 534-6290 Fax: +1 858 534 1625
*Email addresses:* mgoldbaum@ucsd.edu (Michael H. Goldbaum), nhatami@ucsd.edu (Nima Hatami)
*URL:* nimahatami.googlepages.com (Nima Hatami)

Color fundus images currently comprise the largest section of the automatic retinal image analysis literature. Color photography of the ocular fundus has the advantage of making the image available to be examined by a specialist in real time or at another location or time, as well as providing photo documentation for future reference. Modern fundus photographs generally span 30° to 60° view of the ocular fundus.

From the medical image analysis prospective, the presence of inter-and intra-image contrast, luminosity and color variability is a challenge; whereas, the difficulty of detecting differences of arteries and veins in the periphery of the retina is another. Even after image contrast and luminosity normalization, the artery to vein ratio (A/V) may be recognized reasonably well only in an area around the optic disc or the center of the ocular fundus. In the periphery of the image (far from the optic disc) vessels become thinner making the separation of arteries and veins more difficult. Moreover, even around the optic disc only vessels close to each other have been reliably recognized as arteries or veins by direct comparison; whereas, vessels far apart from each other can be easily misclassified.

We propose a new method for extracting robust features for distinguishing retinal arteries and veins. In addition, we provide a comprehensive review on automatic classification of retinal arteries and veins (AV) and measurement of the arteriolar-to-venular diameter ratio (AVR). The proposed method for extracting features is based on Local Binary Pattern (LBP), which provides information from vessel shape and texture. This method validated on retinal fundus images from the UCSD STARE dataset. Comparisons with other classification methods using different learning algorithms and features show a boost in the proposed AV classification task.

This paper is organized as follows: section 2 is dedicated to review of state-of-the art methods in AV classification and AVR measurements. Section 3 presents the proposed new method using LBP for AV classification, followed by an introduction on the LBP method. Experimental results and analyses presented and discussed in section 4 and 5. Section 6 concludes the paper by discussing the value of this new method and by giving a few suggestions for the future research work.

2. Retinal Artery vs Vein Classification

There are diverse methods for automatic AV classification in the fundus image analyses literature. Many of them have tried to simplify the problem by choosing only main vessels around the Optic Nerve Head (ONH). This selection was performed to restrict the analysis to main vessels and avoid the possibly confusing information coming from small arterioles and venules. If the goal is to determine the caliber of the arteries and veins crossing a circle just outside the optic nerve head, determination of the vessel just beyond the ONH border would be sufficient. Abnormalities and changes in vessel structure can affect arteries and veins very differently. Some lesions are peculiar to only one type of vessels, e.g. focal narrowing for arteries, beading for veins. Moreover, one of the early signs of retinopathy is the so-called generalized arteriolar narrowing, in which a ratio between arteries and veins diameters decreases. Sometimes at the crossing of retinal arteries and veins, the vessel can appear to narrow focally or suddenly narrow at a vessel crossing (banking). We often want to know the caliber of the vessel to determine dilation or constriction throughout the ocular fundus, not just in the peripapillary region. The approach to these various analyses of retinal vessels relies on the correct classification of arterial and venous segments.

There are mainly four different characteristics that have been used in the literature to distinguish between retinal arteries and veins: (I)arteries differ in color from veins; (II) arteries are thinner than adjacent veins; (III) the central reflex is wider in arteries than in comparably sized veins, and (IV) arteries and veins usually alternate near the optic disk before branching out, though there can be exceptions. For example, if a branch of the central retinal artery branches again within the border of the optic nerve head, there could be two adjacent arteries outside the border of the ONH.

A Bayesian pixel classifier was proposed by Sim and de Ves in 2001 [1] to distinguish between background, fovea, veins and arteries. This method provides a good segmentation of the main vessels, but it is less effective on the little branches.

Li et al have proposed a piecewise Gaussian model to describe the intensity distribution of vessel profile [2]. They consider the central line reflex of a vessel as a main characteristic with which to distinguish arteries from veins. Using parameters in the best-fitted Piecewise Gaussian model as feature set, the minimum Mahalanobis distance classifier is then employed in the classification of vessel segments.

At the same time as the report by Li et al, Grisan and Ruggeri [5] were applying a *divide and conquer* approach that partitioned a concentric zone around the ONH into quadrants. Each region must contain one of the main arcs of the A/V network: superior-temporal (ST), inferior-temporal (IT), superior-nasal (SN) and inferior-nasal (IN). For this purpose, the optic disc and its approximate diameter needed to be identified first. The fact that the main vessels emerge at the optic disc and then follow a double-parabolic path by branching and thinning is the key assumption for their AV classification technique. For each vessel segment, the variance of red values and the mean of hue values are used together as feature set. First, it classifies the vessels only in a well-defined concentric zone around the optic disc. Then, by using the vessel structure reconstructed by tracking techniques, it would propagate this classification outside this zone, where little color/texture information is available to discriminate arteries from veins. The AV classification algorithm would not be designed considering all the vessels together in the zone, but rather partitioning the area into four quadrants and working separately and locally on each of them.

Later, in 2008 [6], Grisan and Ruggeri simplified their original algorithm with a simple parameter they called *red contrast*, defined as the ratio between the peak of the central line intensity and the largest value between intensities at two vessel edges. A simple threshold is used to decide between arteries or veins.

Narasimha et al. [3] used dual-wavelength imaging to provide both structural and functional features that can be exploited for vessel identification. They presented an automated method to classify arteries and veins in dual-wavelength retinal fundus images recorded at 570 and 600 nm. Using the relative strength of the vessel central reflex for arteries compared to veins and inspired by Lis work [2], they used a dual-Gaussian process to model the cross-sectional intensity profile of vessels. The model parameters are considered as the structural feature. The functional feature exploits the difference in the reflection spectral curves of oxyhemoglobin, which is in high concentration in arteries, and deoxyhemoglobin, which is in high concentration in veins. Therefore, the vessel optical densities ratio (ODR) from a wavelength where the curves differ and one where the curves overlap (isobestic point), ($ODR = OD_{600}/OD_{570}$) as a functional indicator. Finally, the type of the vessel is identified by combining both structural and functional features and using them as classifiers input.

Two feature vector composition methods - profile based and region of interest (ROI) based - have been presented together with two classification methods, Support Vector Machine (SVM)s and Multi-Layer Perceptron (MLP) in [4]. The profile based features are all three color channels after subtraction of the mean value, along each skeleton pixel and the pixels connecting it to the vessel edges orthogonally. The ROI based features are a quadratic region of interest around each skeleton pixel (centered on the center of gravity of the vessel inside the ROI and rotated to align the main axis of the vessel with the horizontal axis). The *combined multiclass principal components analysis*, applied as a dimension-reduction method, is obtained on three sets: 1) all training data samples, 2) artery data samples only, 3) vein data samples only [11]. With hand-segmented vessel data, this approach obtains 95.3% classification rate on the main vessels near the optic disk, given a hand-segmented vessel data, while this value drops by 10% on average, if automatically segmented images are used for the classification. Although, their experiment is limited to the high quality images, and the optic disk is in the center of the image so the main vessels are not so much affected by the shading effect.

A combined clustering and classification approach for separating arteries and veins was presented by Vzquez et al. [8]. The authors compared different feature sets and three classification approaches. The first strategy classified globally all detected vessels applying a clustering algorithm once. The second one divided the retinal image into four quadrants and classified the vessels that belong to the same quadrant independently of the rest of vessels. The third strategy classified the vessels by dividing the retinal image into quadrants that are rotated. Several feature vectors based on the RGB, HSL, and gray scale models were analyzed in order to obtain a precise vessel characterization. The results indicated that the best strategy was the third one, because it minimized the error rate and the number of unclassified vessels; whereas, the most discriminant feature vector was based on the mean or the median of the green component of the RGB color space.

In [7] by Niemeijer et al., intensity and derivative information based features were extracted from each centerline pixel and was used to assign a soft label, indicating the likelihood that it was part of a vein. As all centerline pixels in a connected segment should be the same type, the soft labels aver-aged and assigned to all centerline pixel in the segment. This method only analysed the main vessels origin from the ONH region.

The feature vector described by Rothaus et al. [9] consisted of three color features and five model features (width, contrast, noise ratio, displacement, and modeling error). A K-means clustering algorithm (K=3) was used to distinguish between the veins, arteries and the undefined vessels. A second interpretation step was then applied to assign the meaningful label (artery, vein, and undefined) to the three clusters. The undefined label was assigned to the cluster with the highest average model error. To distinguish the artery and vein cluster, the anatomical description was utilized: A cluster was classified as veins if its corresponding vessels were broader, darker, more reddish, and had a lesser center light reflex than arteries. In 80% of the test images, the classification error was less than 30%.

Zamperini et al. [10] performed several tests investigating various groups of features useful to AV classification. Their results showed that color contrast between vessels and background appear the most important cue for discrimination, but there are vessels that require more features for classification. Vessel position and color variations inside the vessel are useful features; whereas, the vessel width is not. Furthermore, they concluded that the image resolution should be taken into account. High resolution images can introduce noise that reduces color information; the same features computed on subsampled images gave better results, even if an excessive subsampling could remove the information about the central reflex in vessels. The dependency of performance on color confirms the importance of normalizing image resolution in studies involving different resolution fundus cameras.

Some research on AV classification found the central vessel reflex as a key feature, but the central reflex is detectable only on the larger diameter vessels. Moreover, using extra parameters such as using vessel ODs or dual-wavelength are not only available in all cases, but also sometimes can make the decision making process more expensive. Also, many AV classification systems use the features that characterize the color and the color variation in the vessel. However, the absolute color of the blood in the vessels varies between images and across subjects. This variation has several causes. Primarily, the amount of hemoglobin oxygen saturation influences the reflectivity of the blood column, and this difference allows the difference between higher saturation arterial from lower saturation venous blood to be visualized. Next, lens absorption for different wavelengths is influenced by aging and the development of cataract, causing shifts in the spectral distribution of light reflected by blood. Individual difference in pigmentation of the retinal pigment epithelium below the blood vessels also influence the spectrum of reflected light. Finally, across examinations, even from the same subject, differences in flash intensity, flash spectrum, nonlinear optical distortions of the camera, flash artifacts, and focus also cause considerable variability. These factors complicate classification substantially and affect accuracy of the system. Therefore, there is a need for a new method operating that is robust under different circumstances. Table 1 provides a summary of the different approaches, reporting their experimental details.

Table 1: Summary of the state-of-the art methods on AV classification.

| author(s) & year | main idea | dataset characteristics | features | classifier | accuracy |
|---|---|---|---|---|---|
| Simó and Ves Ede 2001 [1] | pixel classification and meta-data based post-processing | fluorescein angiographies, 4 images | pixel intensity and gradient information | Bayesian | Error measures: Ave FP (19.77) and FN (12.6) |
| Li et. al. 2003 [2] | Gaussian model to describe vessel center-line | 505 pre-segmented vessels | Gaussian parameters | Mahalanobis distance | 82.4% arteries and 89% veins |
| Grisan and Ruggeri 2003 [5] | classifying vessels around ONH and tracking them back | 35 fundus image with 45 or 50 field of view | variance of red value and mean of hue | fuzzy clustering | 88% overall |
| Narasimha et. al. 2007 [3] | dual-gaussian to model cross-sectional intensity profile of vessel | 25 dual wavelength images & Gaussian parameters | Gaussian parameters | SVM | 97% arteries and 90% veins |
| Kondermann et. al. 2007 [4] | combining profile-based and ROI-based features | 10132 vessel skeleton pixels from 4 images | RGB values of vessel profile line and ROI | MLP and SVM | 95.32% on vessel pixels |
| Vázquez et al. 2010 [8] | combining clustering and vessel tracking | 58 images centered at the optic disc, VICAVR dataset | RGB, hue, and gray level | rotating quadrants clustering | 90.08% overall |
| Niemeijer et. al. 2009 [7] | experimental investigation of different features/classifiers | 40 fundus images from the DRIVE database | Hue, saturation and intensity | kNN | 88% on centerline pixels |
| Rothaus and Jiang 2011 [9] | assigning vascular area into artery, vein, and undefined clusters and voting as a post-processing | 448 retinal images from MARS data | color and five model features | K-Means clustering | 70% overall |
| Zamperini et. al. 2012 [10] | finding optimal features using spatial location and vessel size | 42 high resolution images, Ninewells Hospital, Dundee | color and vessel position | Linear Bayes | 93.1% overall |

3. Methods

This section describes a novel method for accurate AV classification using the Local Binary Pattern (LBP) features. First a brief introduction of the traditional LBP is given; and in the following the various steps of the proposed LBP-based vessel classification i.e. preprocessing, feature extraction and classification is described in details.

### 3.1. Local Binary Patterns

LBP features, originally proposed by Ojala et al. [12], is an image operator that transforms an image into an array or image of integer labels describing small-scale appearance of the image. These labels or their statistics, most commonly the histogram, are then used for further image analysis. The most widely used versions of the operator are designed for monochrome still images, but this operator has been extended also for color (multi channel) images as well as videos and volumetric data. It is a simple yet efficient local descriptor, and its *Multiscale* [13] version is demonstrated to be rotation-invariant, gray scale-invariant, and multi-resolution. In practice, the LBP operator combines characteristics of statistical and structural texture analysis; it describes the texture with micro-primitives and their statistical placement rules.

The LBP operator has been successfully applied to a variety of uses, such as texture analysis [21], face recognition [15], background modeling [16], gait recognition [18], human activity recognition [19], moving object detection [16], visual speech recognition [17], palm-print recognition [14], texture segmentation [25] and facial expression analysis [20]. LBP has also been effective in medical image analysis tasks, such as texture classification in lung CT images [24], macular pathology diagnosis in retinal OCT images [22], and mammographic mass detection [23].

LBP is a non-parametric kernel which summarizes the local structure around a pixel. For each pixel in an image, a binary code is produced by thresholding its neighborhood (8 pixels when R=1) with the value of the center pixel. A histogram is then constructed to collect the occurrences of different binary patterns representing different types of curved edges, spots, flat areas, etc. The formal LBP equation can be written as:

$$LBP_{P,R} = \sum_{p=0}^{P-1} s(g_p - g_c)2^p \quad s(x) = \begin{cases} 1, & \textit{if } x \geq 0; \\ 0, & \textit{otherwise}. \end{cases} \quad (1)$$

where $g_c$ represents the center pixel and $g_p$ (p=0,...,P-1) denotes its neighbor on a circle of radius $R$, and $P$ is the total number of the neighbors. The neighbors that do not fall in the center of pixels can be estimated by bilinear interpolation. Figure 1 shows how the original LBP is calculated for $LBP_{8,1}$.

The original LBP is based on the assumption that texture has locally two complementary aspects, a pattern and its strength. It is proposed as a two-level version of the texture unit to describe the local textural patterns and considers only the eight nearest neighbors of each pixel. It is rotation variant, but invariant to monotonic changes in gray-scale. There have been major improvements on the traditional LBP, which makes it popular and effective in different application domains. These improvements try to address two main problems i.e. rotations of the input image (Rotation Invariant LBP) and small spatial support area, which only considers *micro* texture information (Multiscale LBP) [13].

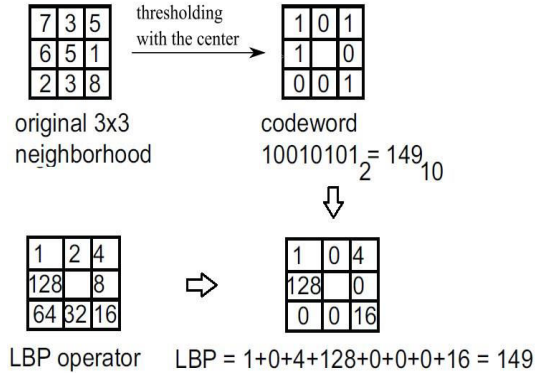

Figure 1: Illustration of the traditional $LBP_{8,1}$

### 3.2. Preprocessing

For retinal vessel localization and segmentation, we used our previously reported method of rotating matched filter with piece-wise thresholding that method that compliments local vessel attributes with region-based attributes of the network structure [28, 27]. A piece of the blood vessel network is hypothesized by probing an area of the Matched Filter Response (MFR) image, iteratively decreasing the threshold. At each iteration, region-based attributes of the piece are tested to consider probe continuation, and ultimately to decide if the piece is vessel. Pixels from probes that are not classified as vessel are recycled for further probing. The strength of this approach is that individual pixel labels are decided using local and region-based properties. A flowchart for the algorithm is shown in Fig. 2.

To analyze the vessel network, we then applied a morphological operation on the binary image of potential blood vessels to thin objects to lines. After removing spur pixels and the objects smaller than 50 pixels, a skeletonization algorithm was applied to reduce all vessels to a centerline one pixel wide. After the skeletonization of the segmented vessels, cross-over and bifurcation points were removed by counting the number of neighbors for all centerline pixels and removing those with more than two neighbors. This is necessary because the vessel width and angle in bifurcations are not well defined and are difficult to measure at cross-over points. This operation subdivides the vascular network into a collection of vessel segments that are individually analyzed.

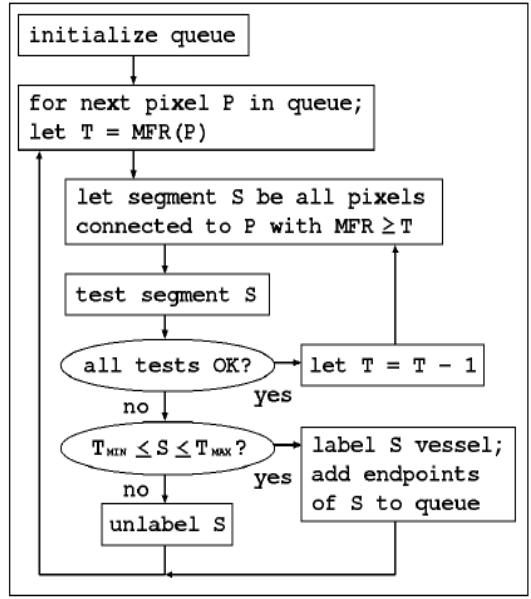

Figure 2: Matched Filter Response-based vessel segmentation.

### 3.3. Feature extraction

robustness to rotations of the input image: As the $LBP_{P,R}$ patterns are obtained by circularly sampling around the center pixel, rotation of the input image has two effects: each local neighborhood is rotated into other pixel locations, and, within each neighborhood, the sampling points on the circle surrounding the center point are rotated into a different orientation. In order to have a rotation-invariant transformation, the first circular symmetric neighborhood is defined as shown in Fig. 3. The pixels that do not exactly match the grid are obtained by interpolation. Fig. 4 shows how the rotation of neighborhood without any pattern change can result in different scores in the traditional LBP. And finally in order to overcome to this change, rotation invariance can be achieved by defining:

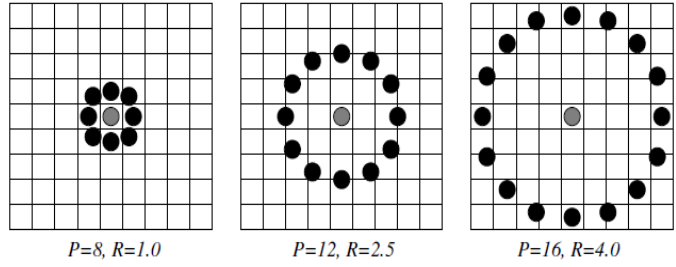

Figure 3: Circularly symmetric neighborhood. Samples that do not exactly match the pixel grid are interpolated.

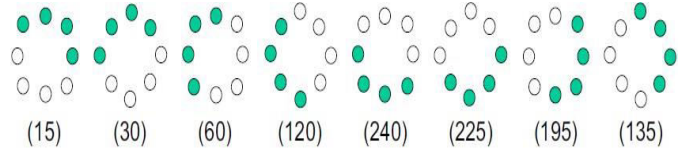

Figure 4: Spatial rotation of the binary pattern changes the LBP code.

$$LBP^{ri}_{P,R} = min\{ROR(LBP_{P,R}, i) \mid i = 0, ..., P - 1\} \quad (2)$$

where $ROR(x, i)$ denotes the circular bitwise right rotation of bit sequence $x$ by $i$ steps. Taking the Fig. 4 as an example, $LBP^{ri}_{8,1} = 15$.

Another significant limitation of the original LBP is its small spatial support area. A straightforward way of enlarging the spatial support area is to combine the information provided by $N$ LBP operators with varying $P$ and $R$ values. This way, each pixel in an image gets $N$ different LBP scores. The most accurate information would be obtained by using the joint distribution of these codes. However, such a distribution would be overwhelmingly sparse with any reasonable image size. The aggregate dissimilarity between a sample and a model can be calculated as a sum of the dissimilarities between the marginal distributions

$$L_N = \sum_{n=1}^{N} L(S^n, M^n) \quad (3)$$

where $S^n$ and $M^n$ correspond to the sample and model distributions extracted by the $n$th operator. The chi square distance or histogram intersection can also be used instead of the log-likelihood measure, $L$. Even though the LBP codes at different radii are not statistically independent in the typical case, using multi-resolution analysis often enhances the discriminative power of the resulting features. With most applications, this straightforward way of building a multi-scale LBP operator has resulted in good accuracy. Fig. 5 shows two version of the multiscale LBP i.e. Multi-Scale Spatial Pyramid (MSSP) and Spatial Pyramid (SP). In order to take advantage of both versions, in this paper we have used Multiscale Rotation Invariant LBP (MS-RI LBP). For further simplicity, figure 6 demonstrates a 2-level LPB operation on an instance fundus image.

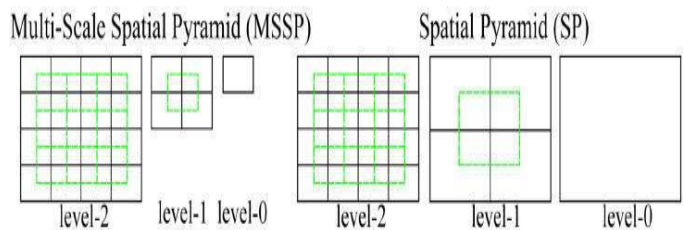

Figure 5: 3-level MSSP and SP.

*3.4. Classification*

This section presents a brief review of Multiple Classifier Systems (MCSs) - used in our experiments - also known in the literature as ensemble learning, committee machine, or combining classifiers. The ultimate goal in pattern recognition is to achieve the best possible classification performance for a given feature set. A promising approach towards this goal consists of combining classifiers, since single classifiers are typically less able to properly handle the complexity of difficult problems. It has been demonstrated that combining a set of independent classifiers with acceptable accuracy leads to better performance [30, 31], provided that the diversity among accurate base classifiers in an ensemble system is enforced in some way.

In order to determine the optimal classification algorithm, both single-classifier techniques (i.e. Bayesian network, Naive Bayes, LibSVM, MLP, Cart and Random tree) and ensemble or multiple classifier (i.e. Adaboost, Bagging, Random Committee, Random subspace, Rotation forest and Majority Voting) have been applied to AV classification problem. A detailed description of the selected methods with a good performance on AV classification is given in [30, 31].

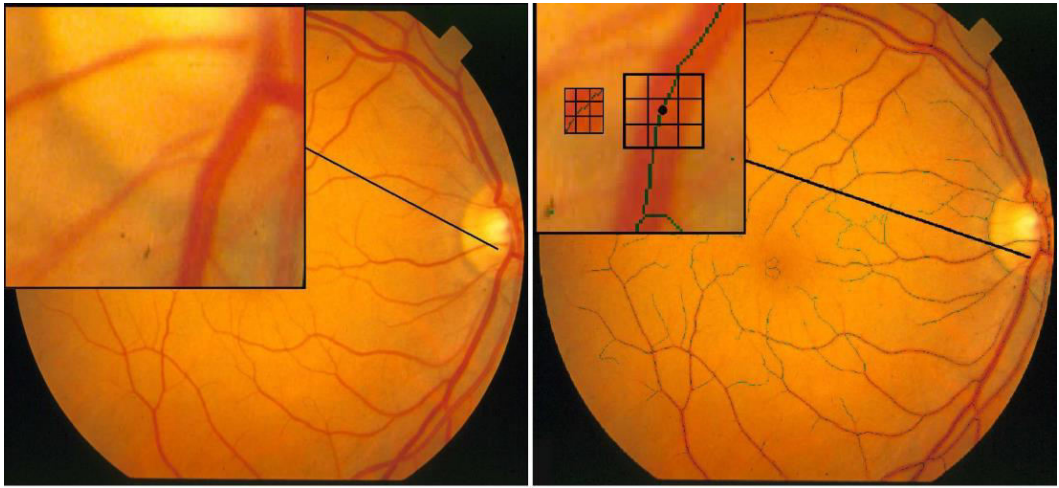

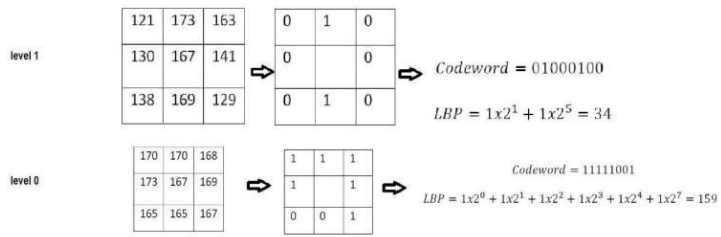

Figure 6: A 2-level LBP-based feature extraction for AV classification. Top-left: an instance fundus image; top-right: two LBP windows on a vessel centerline point; bottom: LBP operation on the gray-scale values.

*3.5. Experimental setup*

To run the experiments, we used images from the STARE (STructured Analysis of the Retina) project [27, 28, 29]. The project was conceived and conducted by one of the authors (MHG) in 1975 and funded by the U.S. National Institutes of Health (NIH). The aim of STARE was to segment, localize, and classify objects of interest in images of the ocular fundus and to infer the diagnosis of the image. Images and clinical data were accumulated from the UCSD based Shiley Eye Center and the Veterans Administration Medical Center in San Diego.

Twenty retinal fundus images were selected for validating the proposed method. Each image was digitized to produce a 1200 × 1400 pixel image, 24-bits per pixel (standard RGB). All ten images contain abnormalities that obscure or confuse the blood vessel appearance. This selection was made for two reasons. First, most of the referenced methods have only been demonstrated upon normal vessel appearances, which are easier to discern. Second, some level of success with abnormal vessel appearances must be established to recommend clinical usage. Each image was carefully hand-labeled by two retina experts to produce a ground truth classification of vessels to serve as a gold standard. After the preprocessing and vessel segmentation, we broke the vessel network from the crossing points into vessel parts (total number of the 614 parts for ten images to use in train/test steps). Training a classifier with the training set, then each instance from the test set sent for the classification and labeling.

In order to compare the proposed method with the state-of-the art and the baseline methods, the popular WEKA (Waikato Environment for Knowledge Analysis) software [26] from the University of Waikato, New Zealand have used. Weka is a collection of machine learning algorithms for data mining tasks. It contains variety of tools for feature extration/selection, classification/clustering, evaluation and visualization. The classification performance assessed by the 10-fold cross-validation which is proven to provide a realistic generalization accuracy for unseen data. The accuracy of the classification models obtained by two main measurements commonly used in the machine learning and pattern recognition literature: recognition rate and the area under the ROC curve (AUC). The area under the ROC curve (AUC) is a very widely used measure of supervised classification performance. It has the appealing property of being objective, requiring no subjective input from the user, and giving a single-value result. The recognition rate is also defined as follows:

$$\mathit{recognitionrate} = \frac{TP + TN}{TP + FP + TN + FN} \times 100 \qquad (4)$$

4. Results

Table 1 and 2 compare the recognition rate and AUC of different classifiers using different feature sets extracted from the vessels. At each row and for each classifier, the best performance obtained by a feature extraction method is highlighted. Average performances and their ranking is also given at the end of each table for ease of comparison. The results of the comparison shows how the proposed method outperforms on AV classification task.

Figure 9 gives an impression about importance of each feature set regardless of the learning algorithm chosen for the classifier (classifier independent analysis). For this purpose, we have measured the recognition rates of different feature sets averaging for all classifiers from table 1. Each bar in the table represent mean and standard deviation of a column from table 1. The big distance of the last two bars indicates the clear winners between different features of vessel parts. It is worth noting that the rest of the feature sets have almost the same performance and how small is their provided information, which makes the classification task for a learner more difficult. In the other hand, the information extracted by LBP and MS-RI LBP from the arteries and veins vessels is so discriminative that no matter which classifier used, it performs better.

In order to evaluate the learning algorithms on AV classification tasks and determine which classifier performs better on the task regardless of selected feature (feature independent analysis), Figure 10 depicts the bar graph of Mean and standard deviation of the recognition rates for different machine learning algorithms. Although there is no clear winner, bagging, random subspace, rotation forest and cart classifiers seem to perform better no matter which feature extraction method has applied. Adaboost and voting techniques also have a lower standard deviation, which demonstrates their stability on the different conditions.

According to the tables 1 and 2 five best AV classification models are as follows: random forest using MS-RI LBP, rotation forest using MS-RI LBP, random tree using MS-RI LBP, random committee using MS-RI LBP, random forest using MS-RI LBP features. Beside the fact that all of the five best models use MS-RI LBP features which proves the efficiency of the proposed method, all of them are also from *classifier ensemble* category. Taking into account the complexity of the AV classification task, it shows how classifiers ensemble could improve the performance of the classification compared to single learners.

Table 1: Recognition rate (%) of the different classifiers using the different features.

| features ▲ classifier ▼ | original RGB | PCA | ICA | Rand. Proj. | wavelet | cluster mem. | Rand. subset | original LBP | MS-RI LBP |
|---|---|---|---|---|---|---|---|---|---|
| Bayesian net. | 51.6 | 60.4 | 59.7 | 60.4 | 50.3 | 60.0 | 52.4 | 60.2 | 66.9 |
| Naive Bayes | 54.8 | 47.9 | 54.9 | 55.5 | 53.0 | 55.3 | 54.8 | 55.0 | 59.6 |
| Lib SVM | 58.4 | 57.1 | 58.0 | 59.2 | 54.8 | 59.9 | 58.8 | 65.5 | 80.3 |
| MLP | 56.3 | 57.4 | 58.0 | 59.9 | 55.5 | 60.3 | 59.0 | 63.7 | 77.4 |
| Adaboost | 58.6 | 58.7 | 59.8 | 59.4 | 57.6 | 59.7 | 59.4 | 65.5 | 61.4 |
| Bagging | 60.0 | 60.2 | 61.8 | 61.7 | 58.1 | 58.0 | 60.5 | 78.0 | 86.7 |
| Rand. Committee | 54.3 | 53.4 | 57.7 | 60.7 | 57.9 | 59.8 | 53.7 | 81.4 | 90.7 |
| Rand. subspace | 59.4 | 59.9 | 61.6 | 60.4 | 59.9 | 61.0 | 60.4 | 77.3 | 89.8 |
| Rot. forest | 57.8 | 59.7 | 62.0 | 60.0 | 57.3 | 61.9 | 57.4 | 81.6 | 90.4 |
| Maj. Voting | 60.4 | 60.4 | 61.3 | 60.4 | 60.4 | 60.4 | 60.4 | 61.8 | 60.9 |
| Cart | 57.9 | 60.4 | 60.0 | 59.4 | 60.4 | 59.1 | 58.9 | 72.4 | 86.6 |
| Rand. Tree | 54.7 | 52.4 | 55.7 | 57.3 | 53.2 | 54.9 | 58.4 | 80.0 | 90.2 |
| Rand. forest | 56.0 | 53.9 | 55.5 | 59.6 | 57.4 | 56.9 | 54.5 | 78.4 | 90.2 |
| Ave. perf. | 56.9 | 57.0 | 58.9 | 59.5 | 56.6 | 59.0 | 57.5 | 70.8 | 79.3 |
| Ranking | 8 | 7 | 5 | 3 | 9 | 4 | 6 | 2 | 1 |

5. Conclusions and Future Work

An automatic retinal artery and vein classification system is proposed using LBP features. Taking advantage of the shape and texture information provided by LBP, the proposed method is not limited to the peripapillary vessels but is also accurate on the vessels elsewhere in the ocular fundus. Furthermore, the proposed method is robust in the presence of illumination change across the retinal fundus images and in low quality samples. Experiments run using nine different feature extraction (commonly used in the state-of-the art AV classification methods) and thirteen machine learning algorithms (both single and ensemble classifiers) in order to provide a broad comparison. Our experiments shows application of simple LBP features on vessel classification task could dramatically improve the performance of any classification system.

Table 2: The area under the ROC curve (AUC) of the different classifiers using the different features.

| features ▲ classifier H | original RGB | PCA | ICA | Rand. Proj. | wavelet | cluster mem. | Rand. subset | original LBP | MS-RI LBP |
|---|---|---|---|---|---|---|---|---|---|
| Bayesian net. | 0.56 | 0.50 | 0.51 | 0.52 | 0.54 | 0.54 | 0.57 | 0.47 | 0.67 |
| Naive Bayes | 0.58 | 0.53 | 0.53 | 0.58 | 0.54 | 0.53 | 0.59 | 0.55 | 0.63 |
| Lib SVM | 0.58 | 0.54 | 0.55 | 0.52 | 0.57 | 0.55 | 0.57 | 0.59 | 0.66 |
| MLP | 0.56 | 0.55 | 0.55 | 0.53 | 0.55 | 0.56 | 0.57 | 0.57 | 0.61 |
| Adaboost | 0.53 | 0.56 | 0.57 | 0.52 | 0.56 | 0.53 | 0.55 | 0.64 | 0.63 |
| Bagging | 0.6 | 0.55 | 0.55 | 0.60 | 0.56 | 0.55 | 0.60 | 0.83 | 0.93 |
| Rand. Committee | 0.55 | 0.51 | 0.53 | 0.60 | 0.55 | 0.56 | 0.54 | 0.91 | 0.97 |
| Rand. subspace | 0.56 | 0.51 | 0.52 | 0.59 | 0.56 | 0.58 | 0.61 | 0.85 | 0.96 |
| Rot. forest | 0.59 | 0.57 | 0.59 | 0.56 | 0.55 | 0.56 | 0.57 | 0.89 | 0.97 |
| Maj. Voting | 0.49 | 0.49 | 0.49 | 0.49 | 0.49 | 0.49 | 0.49 | 0.49 | 0.49 |
| Cart | 0.48 | 0.49 | 0.50 | 0.53 | 0.49 | 0.52 | 0.51 | 0.72 | 0.89 |
| Rand. Tree | 0.53 | 0.50 | 0.52 | 0.55 | 0.51 | 0.54 | 0.57 | 0.78 | 0.90 |
| Rand. forest | 0.53 | 0.52 | 0.55 | 0.60 | 0.57 | 0.56 | 0.57 | 0.87 | 0.97 |
| Ave. AUC | 0.54 | 0.52 | 0.53 | 0.55 | 0.54 | 0.54 | 0.56 | 0.70 | 0.79 |
| Ranking | 5 | 7 | 6 | 4 | 5 | 5 | 3 | 2 | 1 |

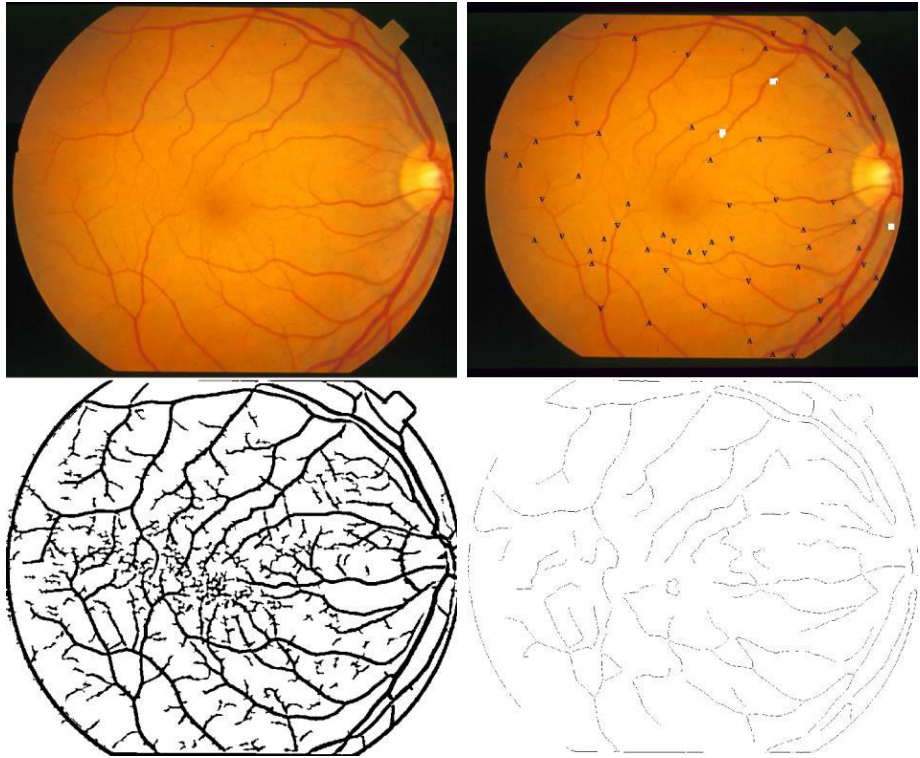

Figure 7: arteries (red color) versus veins (blue color); left: original image, middle: hand-labeled image by the gold standard, right: after vessel segmentation, and bottom-right: vessel center lines.

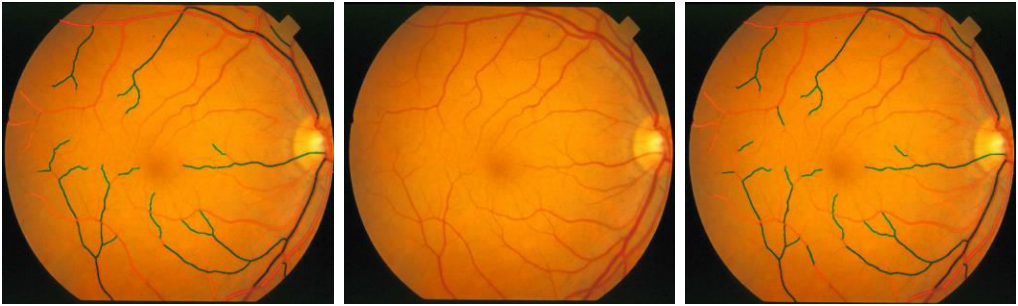

Figure 8: Left: hand-labeled image by the gold standard; middle: a raw image from the dataset; right: results of the proposed automatic system.

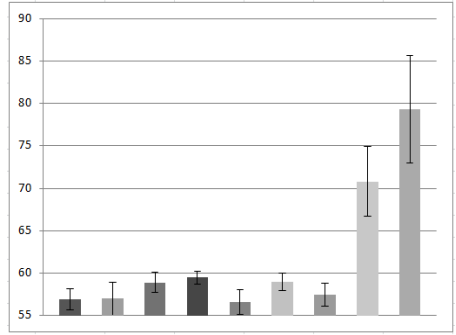

Figure 9: Mean and standard deviation of AV classification (%) for different feature sets (classifier independent). Feature sets (bars) from left to right: original RGB, PCA, ICA, random projection, wavelet, cluster membership, random subset, traditional LBBP and MS-RI LBP.

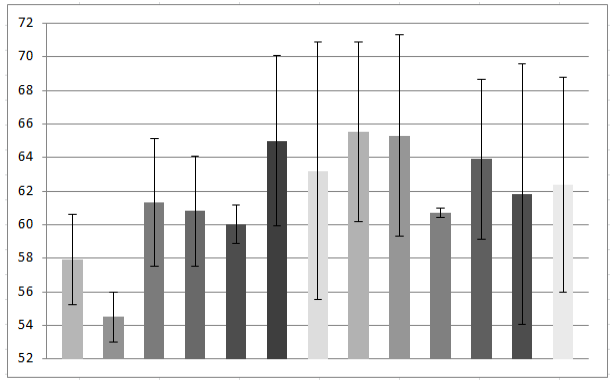

Figure 10: Mean and standard deviation of AV classification (%) for different machine learning algorithms (feature independent). Classifiers (bars) from left to right: bayesian network, naive bayes, libSVM, MLP, adaboost, bagging, random committee, random subspace, rotation forest, majority voting, cart, random tree and random forest.

Several future research directions could be suggested. One of the final goals of this work is the development of a system that computes the arteriolar-to-venular (AVR) diameter ratio automatically. To this end, we need to combine our methodology with an algorithm for the automatic location of optic disc. Moreover, the classification error rates can be reduced by including meta-knowledge such as arteries and vein usually come in pairs and one artery is usually next to two veins and vice versa. Another possible future direction for further improving the AV recognition rate is to combine different features (e.g. ICA) with the LBP features and use it as an input for classifier. Including anatomical characteristics of the vessels and vessel position information in retina could be informative too.

The proposed method may not be limited to retinal vessel analyses and it may be possible to apply it to any vascular system (keeping in mind that each task might require different preprocessing step). This is one of important future lines should be explored by different medical imaging application areas.

*Acknowledgments.*

The authors would like to thank Tzyy-Ping Jung for valuable suggestions that helped improve this paper. This work has been partially supported by EY022039, P30EY022589 National Eye Institute (NEI) Grant, David and Marilyn Dunn Fund and Research to Prevent Blindness.